\pdfoutput=1

\documentclass[11pt]{article}

\usepackage[]{acl}
\usepackage{color}
\usepackage{graphicx}
\usepackage{soul}
\usepackage{booktabs}

\usepackage{tikz}
\usepackage{tikz-cd}
\usepackage{multirow}
\usepackage{subcaption}
\usepackage{amsmath}
\usepackage{amsfonts}
\usepackage{float}
\usetikzlibrary{shapes,decorations,arrows,calc,arrows.meta,fit,positioning}
\tikzset{
    -Latex,auto,node distance =1 cm and 1 cm,semithick,
    state/.style ={ellipse, draw, minimum width = 0.7 cm},
    point/.style = {circle, draw, inner sep=0.04cm,fill,node contents={}},
    bidirected/.style={Latex-Latex,dashed},
    el/.style = {inner sep=2pt, align=left, sloped}
}
\graphicspath{{./images/}}
\usepackage{times}
\usepackage{latexsym}

\usepackage[T1]{fontenc}

\usepackage[utf8]{inputenc}

\usepackage{microtype}

\title{Decomposing Natural Logic Inferences for Neural NLI}
\author{Julia Rozanova$^{1}$,~ Deborah Ferreira$^{1}$,~ Mokanarangan Thayaparan$^{1}$,~ \\
{\bf Marco Valentino$^{1}$ \and Andr\'{e} Freitas$^{1,2}$}\\
Department of Computer Science, University of Manchester, United Kingdom$^{1}$ \\
Idiap Research Institute, Switzerland$^{2}$ \\
\texttt{\{firstname.lastname\}@manchester.ac.uk}}
\begin{document}
\maketitle
\begin{abstract} 
In the interest of interpreting neural NLI models and their reasoning strategies, 
we carry out a systematic probing study which investigates whether these models
capture the crucial semantic features central to natural logic: \emph{monotonicity} and \emph{concept inclusion}.
Correctly identifying valid inferences in \emph{downward-monotone contexts} is a known stumbling block for NLI performance,
subsuming linguistic phenomena such as negation scope and generalized quantifiers.
To understand this difficulty, we emphasize monotonicity as a property of a \emph{context} and examine the extent to which 
models capture relevant monotonicity information in the vector representations which are 
intermediate to their decision making process.
Drawing on the recent advancement of the probing paradigm,
we compare the presence of monotonicity features across various models.
We find that monotonicity information is notably weak in the representations of popular
NLI models which achieve high scores on benchmarks, 
and observe that previous improvements to these models based on fine-tuning strategies 
have introduced stronger monotonicity features together with their improved performance on challenge sets.
\end{abstract}

\section{Introduction}
Large, black box neural models which achieve high scores on benchmark datasets designed for 
testing \emph{natural language understanding} are the subject of much scrutiny 
and investigation.


It is often investigated whether models are able to capture specific semantic phenomena which mimic human reasoning and/or logical formalism, as there is evidence that they sometimes exploit simple heuristics 
and dataset artifacts instead \cite{mccoy, mednli_not_immune}.

In this work, we consider the rigorous setting of \emph{natural logic} \cite{maccartney-manning}.
This is a highly systematic reasoning principle 
relying on only two abstract features, each of which is in itself linguistically complex: \emph{monotonicity} and \emph{concept inclusion word-pair relations}.
It underlies the majority of symbolic/rule-based and hybrid approaches to NLI  and is 
an important baseline reasoning phenomenon to look for in a robust and principled NLI model.

Downward monotone operators such as negation markers and generalized quantifiers result in the kinds of 
natural logic inferences which 
are often known to stump neural NLI models that demonstrate high performance on large benchmark sets such as 
MNLI \cite{mnli}: this has been identified in behavioural studies based on targeted challenge test sets, such as in \citet{yanakaMED} and \citet{geiger}.

In this work, however, we present a \textbf{structural} study: we investigate the extent to which the features relevant for identifying natural logic inferences, especially context monotonicty itself, are captured in the model's internal representations.
To this end, we carry out a systematic \emph{probing} study.

\textbf{Our contributions are may be summarized as follows:}
\begin{enumerate}
    \item We perform a structural investigation as to whether the behaviour of \emph{natural logic} 
    formalisms are mimicked within popular transformer-based NLI models.
    \item For this purpose, we present a joint NLI and semantic probing dataset format (and dataset) which we call NLI-$XY$:
    it is a unique probing dataset in that the probed features relate to the NLI task output in a very systematic way.
    \item We employ thorough probing techniques to determine whether the abstract semantic features of \emph{context monotonicity} and \emph{concept inclusion relations} are captured in the models' internal representations.
    \item We observe that some well-known NLI models demonstrate a 
    systematic failure to model context monotonicity,
    a behaviour we observe to correspond to poor performance on 
    natural logic reasoning in downward-monotone contexts. However, we show that the existing HELP dataset improves this behaviour. 
    \item We support the observations in the probing study with several \emph{qualitative analyses},
    including decomposed error-breakdowns on the \textbf{NLI-XY} dataset, 
    representation visualizations, and evaluations on existing challenge sets.
\end{enumerate}

\section{Related Work}
Natural logic dates back to the formalisms of \citet{sanchez-valencia}, but has been received more recent treatments and reformulations in \citet{maccartney-manning} and \citet{hu-moss}.
Symbolic and hybrid neuro-symbolic implementations of the natural logic paradigm have been explored in \citet{neurallog, hynli, langpro} and \citet{monalog}.

The shortcomings of natural logic handling in various neural NLI models have been shown with several \emph{behavioural} studies, where NLI challenge sets 
exhibiting examples of downward monotone reasoning are used to evaluate performance of models with respect to these
reasoning patterns \cite{mossfragments, yanakaHELP, yanakaMED, goodwin,
geiger}.

In an attempt to better identify linguistic features that neural models manage or fail to capture, 
researchers have employed \emph{probing} strategies: namely, the \emph{diagnostic classification} \cite{alainbengio} 
of auxiliary feature labels from internal model representations. 
Most probing studies in natural language processing focus on the
\emph{syntactic} features captured in transformer-based language 
models \cite{hewitt-manning}, but calls have been made for more sophisticated probing tasks which
rely more on contextual information \cite{pareto}.

In the realm of semantics, probing studies have focused more on \emph{lexical} semantics \cite{vulic}:
word pair relations are central to monotonicity reasoning, and thus form part of our probing study as well,
but the novelty of our work is the task of classifying context monotonicity from intermediate contextual embeddings. 

\section {Problem Formulation: Decomposing Natural Logic}
Natural logic inferences (as formalized in \citealt{sanchez-valencia, maccartney-manning})
are usually described with respect to \emph{substitution} operations.
Certain word substitutions result in 
either forward or reverse entailment, while others result in neither.
This is the basis for a calculus of determining entailment from 
substitution sequences \cite{maccartney-manning, monalog, hu-moss}.

Broadly speaking, we wish to determine whether well-known transformer-based NLI models mimic 
the reasoning strategies of natural logic. 
However, as neural NLI models are black box classifiers that only see a
premise/hypothesis sentence pair as its input, 
it is not immediate how to compare its process to a rule-based system.

To this end, we consider a formulation of natural logic which describes its rules
in terms of concept pair relations and \emph{context monotonicity} (similar to \citealt{rozanovanaloma}).

\subsection{Inferences From Concepts and Contexts}
Consider the following example of a single step natural logic inference, which we will decompose into semantic components relevant to its entailment label: 
\begin{table}[h!]
\resizebox{\columnwidth}{!}{%
\begin{tabular}{@{}lll@{}}
\toprule
                        &                                              & \textbf{NLI Label}          \\
                        \midrule
\textbf{Premise}        & I did not eat any \textbf{fruit} for breakfast.       & \multirow{2}{*}{Entailment} \\
\textbf{Hypothesis}     & I did not eat any \textbf{raspberries} for breakfast. &                            \\
\hline
\end{tabular}%
}
\end{table}

\noindent The hyponym/hypernym pair $($raspberries, fruit$)$ exemplifies a more general relation which we will 
refer to as the \emph{concept inclusion}  
relation $\sqsubset$, (and dually, \emph{reverse concept inclusion} $\sqsupset$).
This mimics the subset relation of the set-based interpretations of the predicates \emph{raspberry} and \emph{fruit}.

In the above example, they occur in a shared \mbox{\textbf{context}}, namely the sentence template 
\begin{quote}
``I did not eat any \rule{1cm}{0.05cm} for breakfast".
\end{quote}

\noindent Such a context may be treated as a term substitution function $f$
$$f: (\mathcal{X}, \sqsubseteq) \to (\mathcal{S}, \Rightarrow)$$
between a set of concepts $\mathcal{X}$ (ordered by the concept inclusion relation) and the set $\mathcal{S}$ of full sentences ordered
by entailment - we demonstrate this substitution in table \ref{tab:tiny_example}.
\begin{table}[h!]
    \centering
    \resizebox{\columnwidth}{!}{
    \begin{tabular}{c|c}
         $X$ & $f(X)$  \\
         \midrule
         \textbf{raspberries} & I did not eat any \textbf{raspberries} for breakfast  \\
         $\sqsubseteq$ &\\
         \textbf{fruit} & I did not eat any \textbf{fruit} for breakfast  \\
    \end{tabular}
    }%
    \caption{In the example, the substitution function $f$ behaves on the concept inputs as shown.}
    \label{tab:tiny_example}
\end{table}

\subsection{Context Monotonicity}
We say that $f$ is \emph{upward monotone} $(\uparrow)$ if it is order \emph{preserving}, i.e.
$$\forall_{X,Y} (X \sqsubseteq Y ~\text{implies} ~f(X) \Rightarrow f(Y))$$
and that $f$ is \emph{downward monotone} $(\downarrow)$ if it is order \emph{reversing}, i.e.
$$ \forall_{X,Y} (X \sqsupseteq Y ~\text{implies} ~f(X) \Rightarrow f(Y)).$$

Given a natural language context $f$, 
any pair of grammatically valid insertions $(X,Y)$ (e.g. ("raspberries", "fruit")) 
yields a sentence pair $(f(X), f(Y))$.
Treating $f(X)$ as a \emph{premise} sentence and $f(Y)$ as a \emph{hypothesis} sentence,
a trained neural NLI model can provide a classification of whether $f(X)$ entails $f(Y).$

In summary, these two abstract linguistic features, 
\emph{context montonicity} and \emph{concept inclusion relation}, jointly determine the final gold entailment label of this type of NLI example.

\begin{figure}[h!]
    \centering
\resizebox{\columnwidth}{!}{%
    \begin{tikzpicture}
            \node (1) [text width=3.5cm, align=center] at (0,0) {Context Monotonicity \\ mon($f$) $\in \{ \uparrow, \downarrow \}$};
        \node (2) [right = of 1]{};
        \node (3) [text width=4.5cm, align=center, right = of 2]{Concept Relation \\ $ rel(\mathbf{X}, \mathbf{Y})\in \{
        =, \sqsubseteq, \sqsupseteq \} $};
        \node (4) [text width=3.5cm, align=center, below = of 2] {Entailment Label for $(f(\mathbf{X}), f(\mathbf{Y}))$};
        \path (3) edge (4);
        \path (1) edge (4);
    \end{tikzpicture}%
    }
\end{figure}

\section{NLI-XY Dataset} 
We follow this formalism as the basis for the \textbf{NLI-$XY$} dataset.
This is the first probing dataset in NLP 
where the auxiliary labels for intermediate semantic features influence the final task 
label in a rigid and deterministic (yet simple) way, with these features being themselves linguistically complex.
As such, it is a "decomposed" natural logic dataset,
where the positive entailment labels are further enriched with labels for the monotonicity and 
relational properties which gave rise to them.
This allows for informative qualitative and structural analyses into natural 
logic handling strategies in neural NLI models. 

\begin{table}[h!]
\resizebox{\columnwidth}{!}{%
\begin{tabular}{@{}llp{3.2cm}p{3cm}@{}}
\toprule
                        & &                                              & \textbf{Auxilliary Label}   \\ \midrule
\textbf{Context}        & $f$ & I did not eat any \rule{0.5cm}{0.05cm} for breakfast.         & $\downarrow$ (downward monotone)                             \\

\textbf{Insertion Pair} & (X,Y) & (fruit, raspberries)                         & $\sqsupset$ (reverse concept inclusion)                    \\ \midrule
                        &                         &                     & \textbf{NLI Label}          \\ \midrule
\textbf{Premise}  & $f(X)$      & I did not eat any fruit for breakfast.       & \multirow{2}{*}{Entailment} \\

\textbf{Hypothesis} & $f(Y)$ &   I did not eat any raspberries for breakfast. &                            \\
\hline \\
\end{tabular}%
}
\caption{A typical NLI-XY example with labels for context monotonicity, lexical relation and the final entailment label. }
\label{tab:example}
\end{table}

The NLI-$XY$ dataset is comprised of the following:
\begin{enumerate}
	\item A set of \emph{contexts} $f$ with a blank position (indicated with a lower case `$x$' or an underscore), annotated with
		the context monotonicity label. 
	\item A set of \emph{insertion pairs} $(X,Y)$, which are either nouns or noun phrases, annotated with the concept inclusion word-pair relation. 
	\item A derived set of premise and hypothesis pairs $(f(X), f(Y)$ made up of permutations of $(X,Y)$ insertion
		pairs through contexts $f$, controlled for grammaticality as far as possible.
\end{enumerate}


\noindent We present examples of the component parts and their composition in table \ref{tab:example}.The premise/hypothesis pairs may thus be used as input to any NLI model, while the context monotonicity and insertion
relation information can be used as the targets of an auxiliary probing task on top of the model's representations.

We make the NLI-$XY$ dataset and all the experimental code used in this work is publically available 
\footnote{\url{https://github.com/juliarozanova/nli_xy}}.
We constructed the NLI-$XY$ dataset used here as follows:
\paragraph{Context Extraction}
We extract context examples from two NLI datasets which were designed for the behavioural
analysis of NLI model performance on monotonicity reasoning. 
In particular, we use the manually curated evaluation set MED \cite{yanakaMED}
and the automatically generated HELP training set \cite{yanakaHELP}.
By design, as they are collections of NLI examples exhibiting monotonicity reasoning, 
these datasets mostly follow our required $(f(X), f(Y))$ structure, and are labeled as instances of upward or downward monotonicity reasoning 
(although the contexts are not explicitly identified).

We extract the common context $f$ from these examples after manually removing a few
which do not follow this structure (differing, for example, in pronoun number agreement or 
prepositional phrases).
We choose to treat determiners and quantifiers as part of the context, as these are the
kinds of closed-class linguistic operators whose monotonicity profiles we are interested in.
To ensure grammatically valid insertions, we manually identify whether each context as suitable either for 
a singular noun, mass noun or plural noun in the blank/``$x$" position. 

\paragraph{Insertion Pairs}
Our $(X,Y)$ insertion phrase pairs come from two sources:
Firstly, the labeled word pairs from the MoNLI dataset \cite{geiger}, which features only 
single-word noun phrases.
Secondly, we include an additional hand-curated dataset which has a small number of \emph{phrase-pair} examples,
which includes intersective modifiers (e.g. ("brown sugar", "sugar")) and prepositional phrases
(e.g. ("sentence", "sentence about oranges")). 
Several of these examples were drawn from the MED dataset. 
Each word in the pair is labelled as a singular, plural or mass noun, so that they may be permuted through the
contexts in a reasonably grammatical way. 

\paragraph{Premise/Hypothesis Pairs}
Premise/Hypothesis pairs are constructed by permuting insertion pairs through the set of contexts within
the grammatical constraints.
Such a permutation strategy may generate examples which are not consistently \emph{meaningful}, but we see
the monotonicity reasoning pattern as sufficiently rigid and syntactic that it is of interest to observe how models treat less "meaningful" entailment examples that still hold with respect to the natural logic formalism:
for example, "I did not swim in a person'' entails "I did not swim in an Irishman" at a systematic level. 
This does raise a question of whether we do (or even should) observe certain systematic behaviours on out-of-distribution examples: we leave the further investigation of this matter for future work.  

Lastly, we note that the data is split into train, dev and test partitions \emph{before} this permutation occurs, so
that there are \textbf{no shared contexts or insertion pairs} between the different data partitions, in an attempt to avoid overlap issues such as those discussed in \cite{question-overlap}.
The full dataset statistics are reported in table \ref{tab:dataset}.

\begin{table}[h]
\centering
\resizebox{0.8\columnwidth}{!}{%
\begin{tabular}{@{}lllll@{}}
\toprule
                   &                         & \multicolumn{3}{c}{\textbf{Context Monotonicity}} \\
\textbf{Partition} & \textbf{(X,Y) Relation} & Up $\uparrow$ & Down $\downarrow$ & Total         \\ \midrule
train              & $\sqsubseteq$           & 671           & 543               & 1214          \\
                   & $\sqsupset$             & 671           & 543               & 1214          \\
                   & None                    & 244           & 222               & 466           \\
                   & Total                   & 1586          & 1308              & \textbf{2894} \\ \midrule
dev                & $\sqsubseteq$           & 598           & 389               & 987           \\
                   & $\sqsupset$             & 598           & 389               & 987           \\
                   & None                    & 220           & 242               & 462           \\
                   & Total                   & 1416          & 1020              & \textbf{2436} \\ \midrule
test               & $\sqsubseteq$           & 1103          & 1066              & 2169          \\
                   & $\sqsupset$             & 1103          & 1066              & 2169          \\
                   & None                    & 502           & 516               & 1018          \\
                   & Total                   & 2708          & 2648              & \textbf{5356} \\
		   \bottomrule \\
\end{tabular}%
}
\caption{Dataset statistics for the NLI-XY dataset. We employ an \textbf{aggressive} $30,20,50$ 
	train-dev-test split for a more impactful probing result, as probing is meant to demonstrate the \emph{ease of extraction} of features. In particular, higher test accuracy with a smaller training set is a more convincing probing result than one with a large training set and small test set.} 
\label{tab:dataset}
\end{table}

\section{Experimental Setup}
Our experiments are designed to investigate the following questions:
Firstly, how do NLI models compare in their learned encoding of context monotonicity and lexical relational features?
Secondly, if a model successfully captures these features, 
to what extent do they correspond with the model's predicted entailment label?  
We investigate these questions with a detailed probing study and a supporting 
qualitative analysis, using decomposed error break-downs and representation visualization. 

\subsection{Model Choices}
We consider a selection of neural NLI models based on BERT-like transformer language models (such as
BERT \cite{bert}, RoBERTa \cite{roberta} and BART \cite{bart}) which are fine-tuned on one of two benchmark training sets:
either SNLI \cite{snli} or MNLI \cite{mnli}. 
Of particular interest, however, is the case where these models are trained on an additional dataset (the HELP dataset 
from \cite{yanakaHELP})
which was designed for improving the overall balance of upward and downward monotone contexts in NLI training data.
We use our own random $50-30-20$ train-dev-test split of the HELP dataset (ensuring unique contexts in every split), 
so that there is no overlap of contexts between the fine-tuning data and the few HELP-test examples we used as part 
of our NLI-XY dataset\footnote{We use the \emph{transformers} library \cite{transformers} and their available pretrained models for this work.}. 

\begin{figure*}[ht!]
    \centering
    \includegraphics[width=1.1\textwidth]{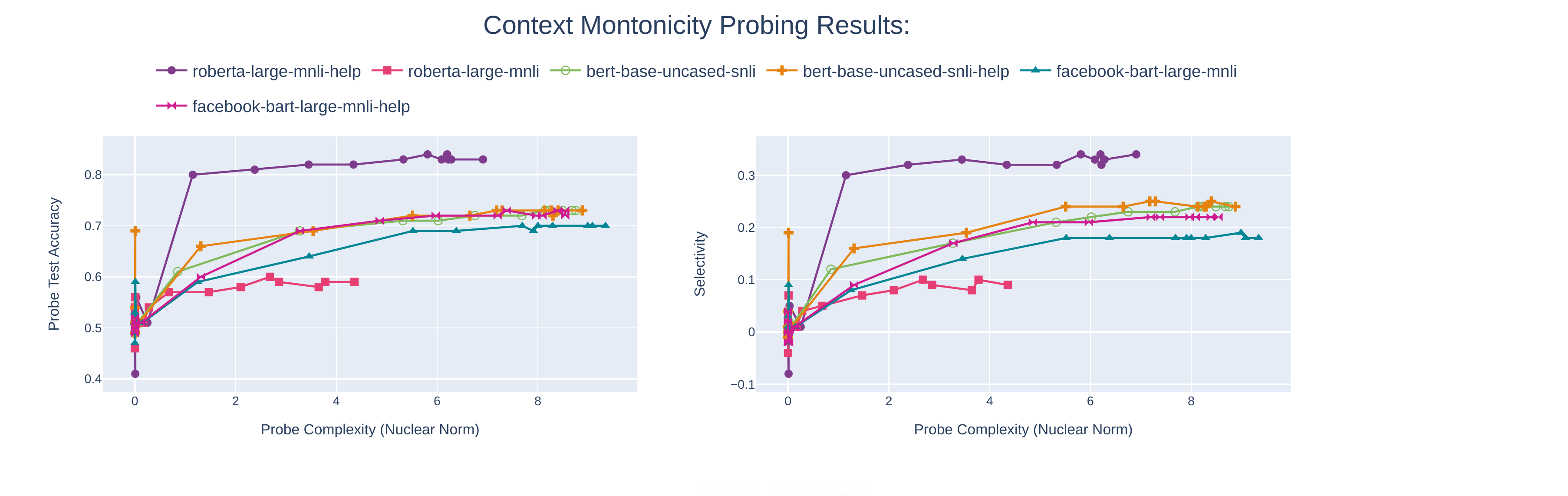}
    \includegraphics[width=\textwidth]{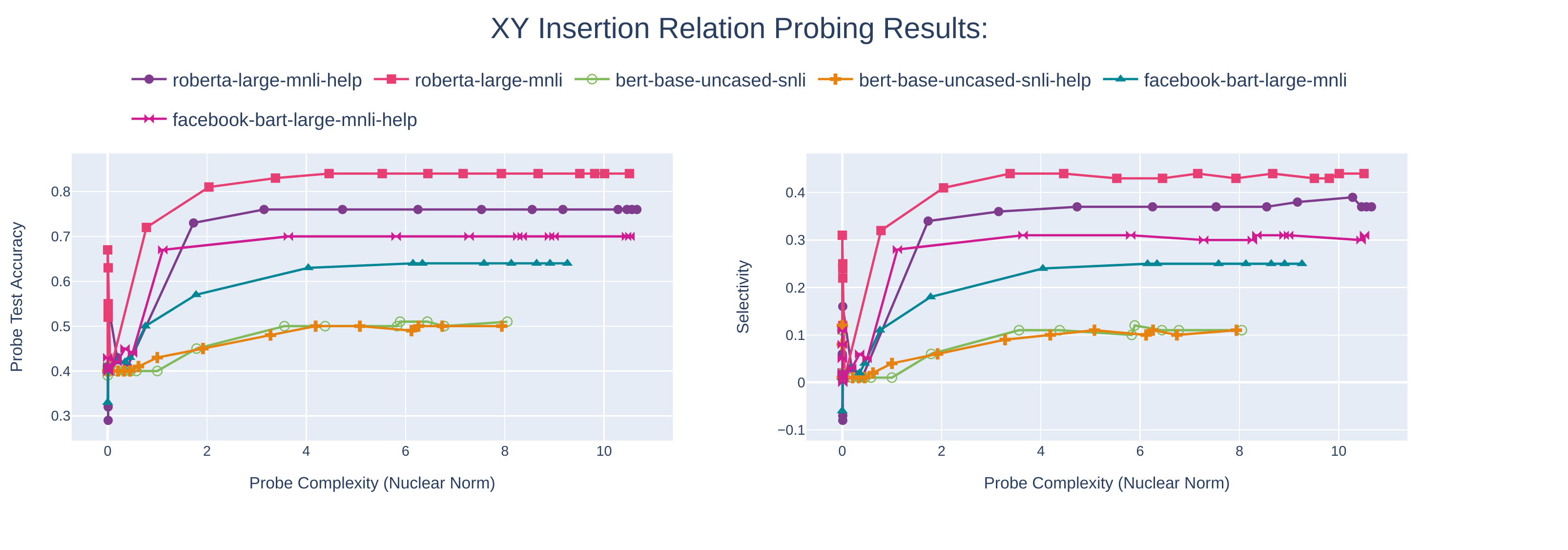}
    \caption{Linear probing results for all examined models.}
    \label{fig:probing}
\end{figure*}

\subsection{Probing Tasks}
The NLI-XY dataset is equipped with two auxiliary feature labels which are the targets 
of the probing task: 
context monotonicity and the relation of the $(X,Y)$ word pair (referred to as concept inclusion relation or lexical relation). We now describe the details of the intermediate representations we choose as inputs to the probing tasks:

\paragraph{Target Representation}
The standard practice for word-pair relation classification tasks is to concatenate the contextual representation vectors for the $(X,Y)$ word pair (taking the mean vector for multi-token words). 
We argue that this is a good representation choice for probing context monotonicity as well: as we are considering transformer-based bidirectional encoder architectures, the context (including the order) of each token in the input sequence informs the representation of each token in the final layer.
As such, we propose that since contextual information is implicitly encoded, it is feasible to expect that a token's vector representation may encode contextual features such as context monotonicity. 
As both the $X$ and the $Y$ word occur in the same respective context, we are comfortable probing the concatenated $(X,Y)$ representation for contextual features, and note that it allows for easy comparison with the word pair relation probing results.

\paragraph{Probing Methodology}
For each auxiliary classification task, we use simple linear models as probes. 
We train 20 probes of varying complexities using the \emph{probe-ably} framework \cite{probe-ably}.

The complexities are represented and controlled as follows: 
For linear models, we follow \citet{pareto} in using the nuclear norm of the linear transformation matrix as the approximate measure of complexity, as it is a continuous approximation of the transformation matrix rank. 
Naively, a strong accuracy on the probing test set may be understood to indicate strong presence
of the target features within the learned representations, 
but there has been much discussion about whether this evidence is compelling on its own.
In fact, certain probing experiments have found the same accuracy scores for 
random representations \cite{zhangbowman}, indicating that high accuracy scores 
are meaningless in isolation.   
\citet{hewitt-liang} describe this as a dichotomy between the representation's encoding of 
the target features and the probe's capacity for \emph{memorization}, and propose the use of 
the \emph{selectivity} measure to always place the probe accuracy in the context of a controlled probing task with shuffled labels on the same vector representations.
For each fully trained probe, we report both the test accuracy
and the \emph{selectivity} measure: tracking the selectivity ensures that we are not using a probe that is
complex enough to be \emph{overly expressive} to the point of 
having the capacity to overfit the 
randomised control training set.
The \emph{selectivity} score is calculated with respect to a \emph{control task}. At its core, this is just
a balanced random relabelling of the auxiliary data, but \citet{hewitt-liang} advocate for more targeted
control tasks with respect to the features in question and a hypothesis about the model's 
possible capacity for \emph{memorization}.
For the lexical relation classification control task, we assign a shared random label for all identical 
insertion pairs, regardless of context.
Thus, a probe which is expressive enough to "memorize" individual labels of word pairs would attain high accuracy on this control task. Analogously, the context monotonicity classification control tasks assigns shared random labels to identical contexts.

\subsection{NLI Challenge Set Evaluations}
As well as the NLI-$XY$ dataset (which can function as an ordinary NLI evaluation set), 
for completeness we report NLI task evaluation scores on the full MED dataset \cite{yanakaMED},
which was designed as a thorough stress-test of monotonicity reasoning performance. 
Furthermore, we report scores on the HELP-test set (from the dataset split 
in \citealt{rozanovanaloma}): this data partition was not used in the fine-tuning of
models on HELP, but we include the test scores here for insight.

\subsection{Decomposed Error Analysis}
The compositional structure and auxiliary labels in the NLI-XY dataset
allow for qualitative analysis which may enrich the observations.
To this end, we construct decomposed error analysis heatmaps which indicate
whether a given premise-hypothesis data point $(f(X),f(Y))$ 
is correctly classified by an entailment model. These are structured with individual $(X,Y)$ insertion pairs on the vertical axis and contexts on the horizontal axis.
For brevity (and because this is representative of our observations),
we include only the error breakdowns for the two sublasses of the 
positive entailment label: where the context monotonicity is upward and lexical relation 
is forward incusion, and where the context monotonicity is downward and the lexical relation
is reverse inclusion. 

\begin{table*}[ht!]
\centering
\resizebox{0.9\textwidth}{!}{
\begin{tabular}{llllllccc}
\hline
\multicolumn{1}{c}{}                    &                                                                & \multicolumn{1}{c}{} & \multicolumn{2}{c}{{\color[HTML]{000000} \textbf{Feature Probing}}}                                                                                                                      & \multicolumn{1}{c}{\textbf{}} & \multicolumn{3}{c}{\textbf{NLI Monotonicity Challenge Sets}}                                                                                                                                \\ \cline{1-2} \cline{4-5} \cline{7-9} 
\multicolumn{1}{c}{\textbf{NLI Models}} & \begin{tabular}[c]{@{}l@{}}Fine-Tuning \\ Data\end{tabular}    & \multicolumn{1}{c}{} & \multicolumn{1}{c}{\begin{tabular}[c]{@{}c@{}}Context \\ Monotonicity \\ (\%*)\end{tabular}} & \multicolumn{1}{c}{\begin{tabular}[c]{@{}c@{}}XY Insertion \\ Relation \\ (\%*)\end{tabular}} & \multicolumn{1}{c}{}          & \begin{tabular}[c]{@{}c@{}}HELP-Test \\ (\%)\end{tabular} & \begin{tabular}[c]{@{}c@{}}MED \\ (\%)\end{tabular} & \multicolumn{1}{l}{\begin{tabular}[c]{@{}l@{}}NLI-XY\\ (\%)\end{tabular}} \\ \cline{1-2} \cline{4-5} \cline{7-9} 
roberta-large-mnli                      & -                                                              &                      & 59.00                                                                                         & 84.00                                                                                      &                               & 36.69                                                     & 46.10                                               & 59.01                                                                     \\
roberta-large-mnli                      & HELP                                                           &                      & 84.00                                                                                         & 76.00                                                                                      &                               & \textbf{97.63}                                            & \textbf{78.22}                                      & \textbf{80.68}                                                            \\
                                        &                                                                &                      &                                                                                              &                                                                                           &                               &                                                           &                                                     &                                                                           \\
facebook/bart-large-mnli                &                                                                &                      & 70.00                                                                                         & 64.00                                                                                      &                               & 43.61                                                     & 46.54                                               & 60.59                                                                     \\
facebook/bart-large-mnli                & HELP                                                           &                      & 73.00                                                                                         & 70.00                                                                                      &                               & 88.99                                                     & 77.16                                               & 79.34                                                                   \\
                                        &                                                                &                      &                                                                                              &                                                                                           &                               & \multicolumn{1}{l}{}                                      & \multicolumn{1}{l}{}                                & \multicolumn{1}{l}{}                                                      \\
bert-base-uncased-snli                  &                                                                &                      & 73.00                                                                                         & 51.00                                                                                      &                               & 63.55                                                     & 49.38                                              & 49.09                                                                     \\
bert-base-uncased-snli                  & HELP                                                           &                      & 73.00                                                                                         & 51.00                                                                                      &                               & 66.80                                                     & 46.13                                              & 44.79                                                                     \\
\hline
\end{tabular}%
}
\caption{Summary NLI challenge test set and probing results for all considered models. $^*$Probing results are summarized with the \emph{accuracy at max selectivity}.}
\label{tab:challenge_comparison}
\end{table*}

\section{Results and Discussion}
\subsection{Probing Results}
The results for the linear probing experiments for both the \emph{context monotonicity classification} task and 
the \emph{lexical relation} classification task may be found in figure \ref{fig:probing}, 
with a summary score of accuracy at maximum selectivity visible in table \ref{tab:challenge_comparison}.
The results of the control tasks are taken into account as part of the selectivity measure, 
which is represented on the right hand plot for each experiment. 

It is particularly notable that large models trained only on the MNLI dataset
have inferior performance on context monotonicity classification. 
This corresponds with the further qualitative observations, suggesting that even in some of the most successful transformer-based NLI models, \emph{there is a poor ``understanding" of 
the logical regularities of contexts and how these are altered with downward monotone operators.}

\subsection{Comparison to Challenge Set Performance}
A summary of the probing results (presented as accuracy at maximum selectivity) can be compared with challenge set performance in table \ref{tab:challenge_comparison}.
Evaluation on the challenge test sets is relatively consistent with monotonicity probing performance, in the sense that there is a correspondence between poor/successful modeling of monotonicity features and 
poor/successful performance on a targeted natural logic test set. 
As these challenge sets are focused on testing monotonicity reasoning, this is a result 
which strongly bolsters the suggestion that explicit representation of the context monotonicity feature is crucial, especially for examples involving negation and other downward monotone operators.
Furthermore, we generally confirm previous results that additional fine-tuning on the HELP data set has been
helpful for these specialized test sets, and add to this that it similarly improves the explicit extractability of relevant context montonicity features from the latent vector representations.

\begin{figure*}[htb!]
\label{fig:decomposed}
\centering
\begin{subfigure}{.5\textwidth}
  \centering
  \includegraphics[width=\textwidth]{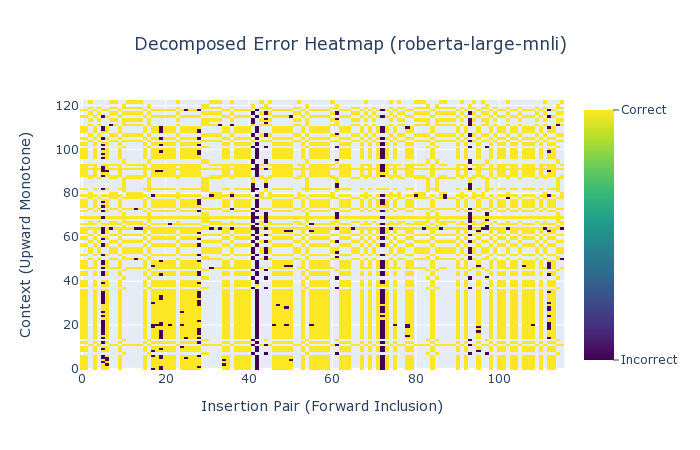}
  \caption{}
\end{subfigure}%
\begin{subfigure}{.5\textwidth}
  \centering
  \includegraphics[width=\textwidth]{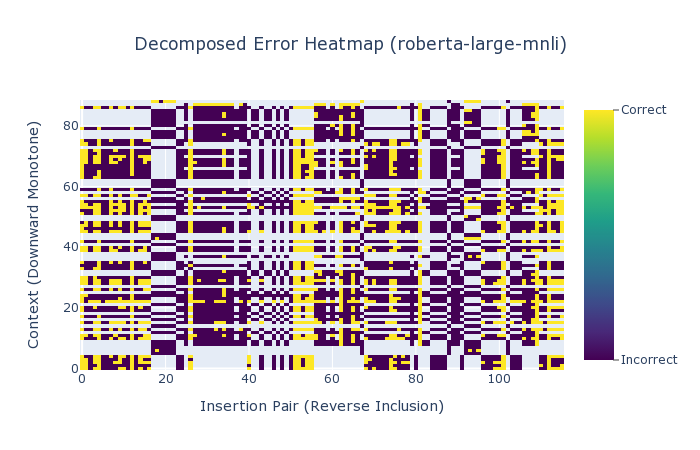}
  \caption{}
\end{subfigure}

\begin{subfigure}{.5\textwidth}
  \centering
  \includegraphics[width=\textwidth]{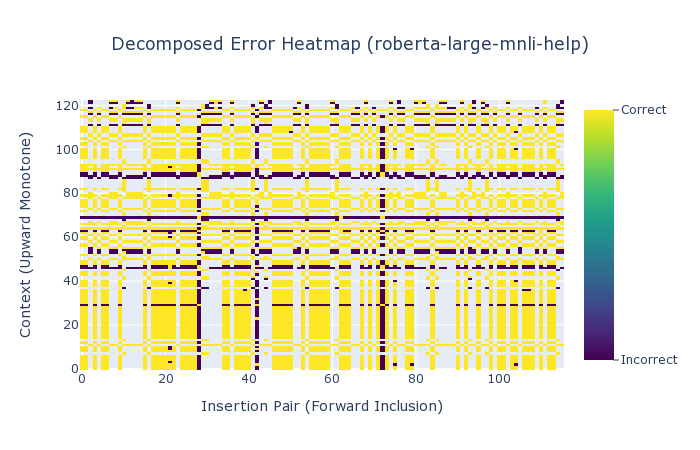}
  \caption{}
\end{subfigure}%
\begin{subfigure}{.5\textwidth}
  \centering
  \includegraphics[width=\textwidth]{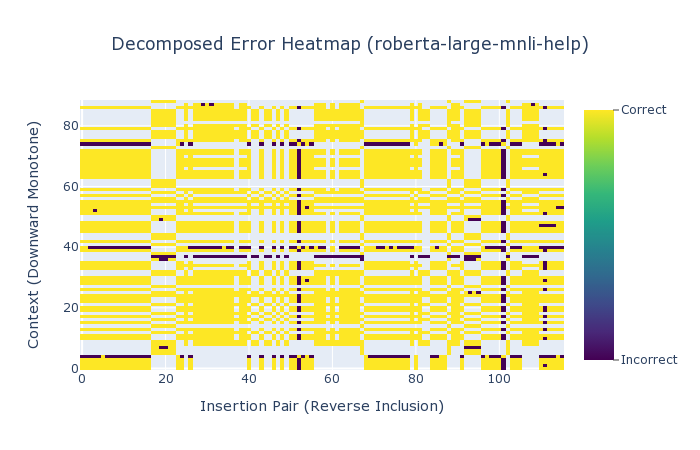}
  \caption{}
\end{subfigure}%
\caption{Decomposed error heat maps for portions of the NLI-XY dataset corresponding to the indicated context monotonicity and insertion relations (blank positions are present as only grammatical insertions were included in the dataset.)}
\end{figure*}
\subsection{Qualitative Analyses}

\begin{figure*}[htb!]
\centering
\begin{subfigure}{.5\textwidth}
  \centering
  \includegraphics[height=5cm]{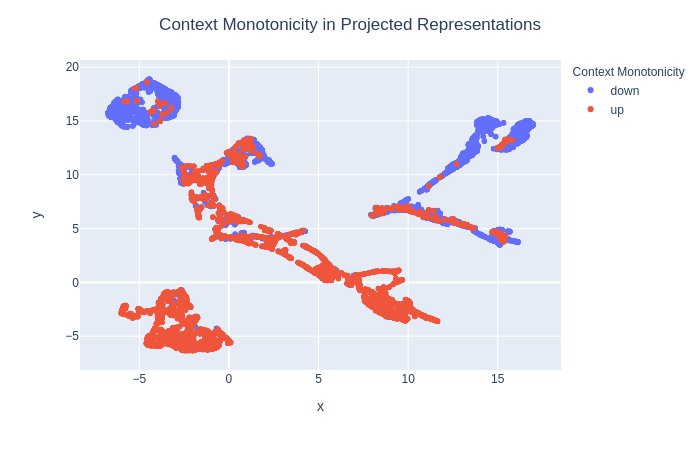}
  \caption{\texttt{roberta-large-mnli-help}}
\end{subfigure}%
\begin{subfigure}{.5\textwidth}
  \centering
  \includegraphics[width=\textwidth]{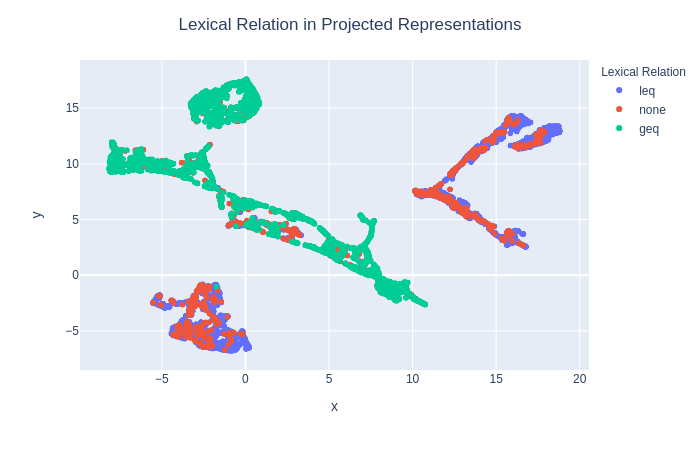}
  \caption{\texttt{roberta-large-mnli-help}}
\end{subfigure}

\begin{subfigure}{.5\textwidth}
  \centering
  \includegraphics[height=5cm]{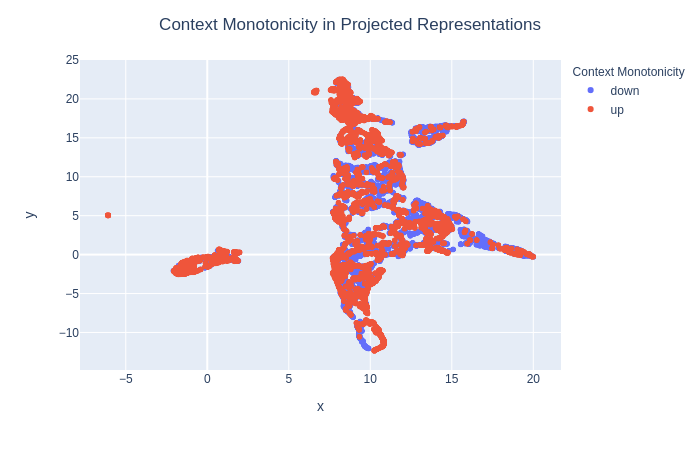}
  \caption{\texttt{roberta-large-mnli}}
\end{subfigure}%
\begin{subfigure}{.5\textwidth}
  \centering
  \includegraphics[width=\textwidth]{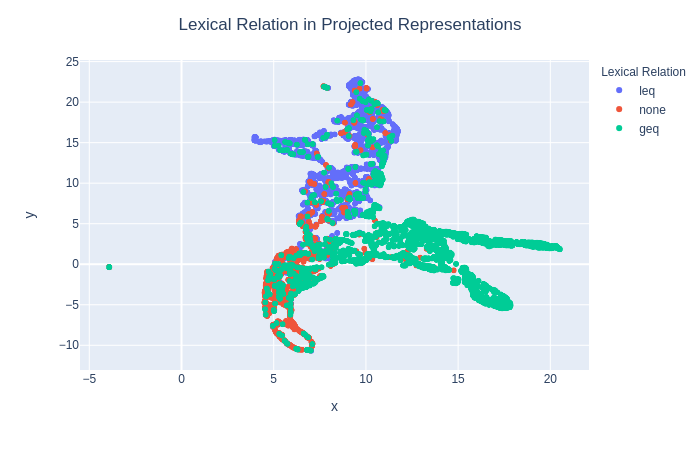}
  \caption{\texttt{roberta-large-mnli}}
\end{subfigure}%
\caption{UMAP projections of selected classification token representations comparing \texttt{roberta-large-mnli} and the improved \texttt{roberta-large-mnli-help}, which shows greater distinction between context monotonicity features.}
\label{fig:umap}
\end{figure*}
\paragraph{Error Break-Downs}
An error heat map according to decomposed context monotonicity and word-pair insertion relation can be seen in figure 2.
We are less concerned with the accuracy score (on NLI challenge sets) of a given model
as with the behavioural \emph{systematicity} visible in the errors, 
as we are not interested in noisy errors which may be due to words or phrases from outside the training domain.
Consistent mis-classification for all examples
derived from a fixed context or insertion pair 
are actually \emph{also} strongly suggestive of a regularity in reasoning. 
The decomposed error analyses paint a striking picture: 
we generally see that models trained on MNLI routinely fail to distinguish between
the expected behaviour of upward and downward monotone contexts, 
despite generally achieving high accuracies on large 
benchmark sets. 
This is in accordance with observations in \citet{yanakaHELP} and \citet{yanakaMED}, where low accuracy on the downward-monotone reasoning sections of challenge sets points to this possibility.
Howver, they show consistently show strong behavioural regularity with respect to concept inclusion. 
Even when the contexts are downward monotone, they still treat them systematically as if they were \emph{upward} monotone, echoing the concept insertion pair relation \emph{only}: they completely fail to discriminate between upward/downward monotone contexts and their opposite behaviours. 
\paragraph{Visualization}
In figure \ref{fig:umap},  
each data point corresponds to an embedded example (contextual $XY$ word pair representation) in the NLI-$XY$ dataset, with the left and right columns colored with the \emph{gold} auxiliary labels for context monotonicity and concept inclusion relations respectively. These illustrate the probing observations: in the well-known \texttt{roberta-large-mnli} model, concept inclusion relation features are distinguishable, whereas context monotonicity is very randomly scattered, with no emergent clustering.  However, the \texttt{roberta-large-mnli-help} model shows an improvement in this behaviour, demonstrating a stronger context monotonicity distinction.
\section{Conclusion}
In summary, the NLI-$XY$ has enabled us to present evidence that explicit context monotonicity feature clustering in neural model representations seems to correspond to better performance on natural logic challenge sets which test downward-monotone reasoning. In particular, many popular models trained on MNLI seem to lack this behaviour, accounting for previous observations that they systematically fail in downward-monotone contexts.

Furthermore, the probes' labels also have some explanatory value: both entailment and non-entailment
labels can each further be broken down into sub-regions. 
This qualifies the classification with the observations that the data point occurs in a 
cluster of examples with a) upward (resp. downward) contexts and b) a forward (resp. backward)
containment relation between the substituted noun phrases.
In this sense, the analyses in this work can thus be interpreted as an explainable ``decomposition" of the treatment natural logic examples in neural models.

\bibliography{anthology,custom}

\begin{thebibliography}{29}
\expandafter\ifx\csname natexlab\endcsname\relax\def\natexlab#1{#1}\fi

\bibitem[{Abzianidze(2017)}]{langpro}
Lasha Abzianidze. 2017.
\newblock \href {https://doi.org/10.18653/v1/D17-2020} {{L}ang{P}ro: Natural
  language theorem prover}.
\newblock In \emph{Proceedings of the 2017 Conference on Empirical Methods in
  Natural Language Processing: System Demonstrations}, pages 115--120,
  Copenhagen, Denmark. Association for Computational Linguistics.

\bibitem[{Alain and Bengio(2018)}]{alainbengio}
Guillaume Alain and Yoshua Bengio. 2018.
\newblock \href {http://arxiv.org/abs/1610.01644} {Understanding intermediate
  layers using linear classifier probes}.

\bibitem[{Bowman et~al.(2015)Bowman, Angeli, Potts, and Manning}]{snli}
Samuel~R Bowman, Gabor Angeli, Christopher Potts, and Christopher~D Manning.
  2015.
\newblock A large annotated corpus for learning natural language inference.
\newblock In \emph{EMNLP}.

\bibitem[{Chen et~al.(2021)Chen, Gao, and Moss}]{neurallog}
Zeming Chen, Qiyue Gao, and Lawrence~S. Moss. 2021.
\newblock \href {http://arxiv.org/abs/2105.14167} {Neurallog: Natural language
  inference with joint neural and logical reasoning}.

\bibitem[{Devlin et~al.(2019)Devlin, Chang, Lee, and Toutanova}]{bert}
Jacob Devlin, Ming-Wei Chang, Kenton Lee, and Kristina Toutanova. 2019.
\newblock \href {https://doi.org/10.18653/v1/N19-1423} {{BERT}: Pre-training of
  deep bidirectional transformers for language understanding}.
\newblock In \emph{Proceedings of the 2019 Conference of the North {A}merican
  Chapter of the Association for Computational Linguistics: Human Language
  Technologies, Volume 1 (Long and Short Papers)}, pages 4171--4186,
  Minneapolis, Minnesota. Association for Computational Linguistics.

\bibitem[{Ferreira et~al.(2021)Ferreira, Rozanova, Thayaparan, Valentino, and
  Freitas}]{probe-ably}
Deborah Ferreira, Julia Rozanova, Mokanarangan Thayaparan, Marco Valentino, and
  Andr{\'e} Freitas. 2021.
\newblock \href {https://doi.org/10.18653/v1/2021.acl-demo.23} {Does my
  representation capture {X}? probe-ably}.
\newblock In \emph{Proceedings of the 59th Annual Meeting of the Association
  for Computational Linguistics and the 11th International Joint Conference on
  Natural Language Processing: System Demonstrations}, pages 194--201, Online.
  Association for Computational Linguistics.

\bibitem[{Geiger et~al.(2020)Geiger, Richardson, and Potts}]{geiger}
Atticus Geiger, Kyle Richardson, and Christopher Potts. 2020.
\newblock \href {https://doi.org/10.18653/v1/2020.blackboxnlp-1.16} {Neural
  natural language inference models partially embed theories of lexical
  entailment and negation}.
\newblock In \emph{Proceedings of the Third BlackboxNLP Workshop on Analyzing
  and Interpreting Neural Networks for NLP}, pages 163--173, Online.
  Association for Computational Linguistics.

\bibitem[{Goodwin et~al.(2020)Goodwin, Sinha, and O{'}Donnell}]{goodwin}
Emily Goodwin, Koustuv Sinha, and Timothy~J. O{'}Donnell. 2020.
\newblock \href {https://doi.org/10.18653/v1/2020.acl-main.177} {Probing
  linguistic systematicity}.
\newblock In \emph{Proceedings of the 58th Annual Meeting of the Association
  for Computational Linguistics}, pages 1958--1969, Online. Association for
  Computational Linguistics.

\bibitem[{Herlihy and Rudinger(2021)}]{mednli_not_immune}
Christine Herlihy and Rachel Rudinger. 2021.
\newblock \href {https://doi.org/10.18653/v1/2021.acl-short.129} {{M}ed{NLI} is
  not immune: {N}atural language inference artifacts in the clinical domain}.
\newblock In \emph{Proceedings of the 59th Annual Meeting of the Association
  for Computational Linguistics and the 11th International Joint Conference on
  Natural Language Processing (Volume 2: Short Papers)}, pages 1020--1027,
  Online. Association for Computational Linguistics.

\bibitem[{Hewitt and Liang(2019)}]{hewitt-liang}
John Hewitt and Percy Liang. 2019.
\newblock \href {https://doi.org/10.18653/v1/D19-1275} {Designing and
  interpreting probes with control tasks}.
\newblock In \emph{Proceedings of the 2019 Conference on Empirical Methods in
  Natural Language Processing and the 9th International Joint Conference on
  Natural Language Processing (EMNLP-IJCNLP)}, pages 2733--2743, Hong Kong,
  China. Association for Computational Linguistics.

\bibitem[{Hewitt and Manning(2019)}]{hewitt-manning}
John Hewitt and Christopher~D. Manning. 2019.
\newblock \href {https://doi.org/10.18653/v1/N19-1419} {{A} structural probe
  for finding syntax in word representations}.
\newblock In \emph{Proceedings of the 2019 Conference of the North {A}merican
  Chapter of the Association for Computational Linguistics: Human Language
  Technologies, Volume 1 (Long and Short Papers)}, pages 4129--4138,
  Minneapolis, Minnesota. Association for Computational Linguistics.

\bibitem[{Hu et~al.(2020)Hu, Chen, Richardson, Mukherjee, Moss, and
  Kuebler}]{monalog}
Hai Hu, Qi~Chen, Kyle Richardson, Atreyee Mukherjee, Lawrence~S. Moss, and
  Sandra Kuebler. 2020.
\newblock \href {https://aclanthology.org/2020.scil-1.40} {{M}ona{L}og: a
  lightweight system for natural language inference based on monotonicity}.
\newblock In \emph{Proceedings of the Society for Computation in Linguistics
  2020}, pages 334--344, New York, New York. Association for Computational
  Linguistics.

\bibitem[{Hu and Moss(2018)}]{hu-moss}
Hai Hu and Larry Moss. 2018.
\newblock \href {https://doi.org/10.18653/v1/S18-2015} {Polarity computations
  in flexible categorial grammar}.
\newblock In \emph{Proceedings of the Seventh Joint Conference on Lexical and
  Computational Semantics}, pages 124--129, New Orleans, Louisiana. Association
  for Computational Linguistics.

\bibitem[{Kalouli et~al.(2020)Kalouli, Crouch, and de~Paiva}]{hynli}
Aikaterini-Lida Kalouli, Richard Crouch, and Valeria de~Paiva. 2020.
\newblock \href {https://doi.org/10.18653/v1/2020.coling-main.459} {Hy-{NLI}: a
  hybrid system for natural language inference}.
\newblock In \emph{Proceedings of the 28th International Conference on
  Computational Linguistics}, pages 5235--5249, Barcelona, Spain (Online).
  International Committee on Computational Linguistics.

\bibitem[{Lewis et~al.(2020)Lewis, Liu, Goyal, Ghazvininejad, Mohamed, Levy,
  Stoyanov, and Zettlemoyer}]{bart}
Mike Lewis, Yinhan Liu, Naman Goyal, Marjan Ghazvininejad, Abdelrahman Mohamed,
  Omer Levy, Veselin Stoyanov, and Luke Zettlemoyer. 2020.
\newblock \href {https://doi.org/10.18653/v1/2020.acl-main.703} {{BART}:
  Denoising sequence-to-sequence pre-training for natural language generation,
  translation, and comprehension}.
\newblock In \emph{Proceedings of the 58th Annual Meeting of the Association
  for Computational Linguistics}, pages 7871--7880, Online. Association for
  Computational Linguistics.

\bibitem[{Lewis et~al.(2021)Lewis, Stenetorp, and Riedel}]{question-overlap}
Patrick Lewis, Pontus Stenetorp, and Sebastian Riedel. 2021.
\newblock \href {https://aclanthology.org/2021.eacl-main.86} {Question and
  answer test-train overlap in open-domain question answering datasets}.
\newblock In \emph{Proceedings of the 16th Conference of the European Chapter
  of the Association for Computational Linguistics: Main Volume}, pages
  1000--1008, Online. Association for Computational Linguistics.

\bibitem[{Liu et~al.(2019)Liu, Ott, Goyal, Du, Joshi, Chen, Levy, Lewis,
  Zettlemoyer, and Stoyanov}]{roberta}
Y.~Liu, Myle Ott, Naman Goyal, Jingfei Du, Mandar Joshi, Danqi Chen, Omer Levy,
  M.~Lewis, Luke Zettlemoyer, and Veselin Stoyanov. 2019.
\newblock Roberta: A robustly optimized bert pretraining approach.
\newblock \emph{ArXiv}, abs/1907.11692.

\bibitem[{MacCartney and Manning(2007)}]{maccartney-manning}
Bill MacCartney and Christopher~D. Manning. 2007.
\newblock \href {https://www.aclweb.org/anthology/W07-1431} {Natural logic for
  textul inference}.
\newblock In \emph{Proceedings of the {ACL}-{PASCAL} Workshop on Textual
  Entailment and Paraphrasing}, pages 193--200, Prague. Association for
  Computational Linguistics.

\bibitem[{McCoy et~al.(2019)McCoy, Pavlick, and Linzen}]{mccoy}
Tom McCoy, Ellie Pavlick, and Tal Linzen. 2019.
\newblock \href {https://doi.org/10.18653/v1/P19-1334} {Right for the wrong
  reasons: Diagnosing syntactic heuristics in natural language inference}.
\newblock In \emph{Proceedings of the 57th Annual Meeting of the Association
  for Computational Linguistics}, pages 3428--3448, Florence, Italy.
  Association for Computational Linguistics.

\bibitem[{Pimentel et~al.(2020)Pimentel, Saphra, Williams, and
  Cotterell}]{pareto}
Tiago Pimentel, Naomi Saphra, Adina Williams, and Ryan Cotterell. 2020.
\newblock \href {https://doi.org/10.18653/v1/2020.emnlp-main.254} {{P}areto
  probing: {T}rading off accuracy for complexity}.
\newblock In \emph{Proceedings of the 2020 Conference on Empirical Methods in
  Natural Language Processing (EMNLP)}, pages 3138--3153, Online. Association
  for Computational Linguistics.

\bibitem[{Richardson et~al.(2019)Richardson, Hu, Moss, and
  Sabharwal}]{mossfragments}
Kyle Richardson, Hai Hu, Lawrence~S. Moss, and Ashish Sabharwal. 2019.
\newblock \href {http://arxiv.org/abs/1909.07521} {Probing natural language
  inference models through semantic fragments}.
\newblock \emph{CoRR}, abs/1909.07521.

\bibitem[{Rozanova et~al.(2021)Rozanova, Ferreira, Thayaparan, Valentino, and
  Freitas}]{rozanovanaloma}
Julia Rozanova, Deborah Ferreira, Mokanarangan Thayaparan, Marco Valentino, and
  André Freitas. 2021.
\newblock \href {http://arxiv.org/abs/2105.08008} {Supporting context
  monotonicity abstractions in neural nli models}.

\bibitem[{Sanchez(1991)}]{sanchez-valencia}
V.~Sanchez. 1991.
\newblock Studies on natural logic and categorial grammar.

\bibitem[{Vuli{\'c} et~al.(2020)Vuli{\'c}, Ponti, Litschko, Glava{\v{s}}, and
  Korhonen}]{vulic}
Ivan Vuli{\'c}, Edoardo~Maria Ponti, Robert Litschko, Goran Glava{\v{s}}, and
  Anna Korhonen. 2020.
\newblock \href {https://doi.org/10.18653/v1/2020.emnlp-main.586} {Probing
  pretrained language models for lexical semantics}.
\newblock In \emph{Proceedings of the 2020 Conference on Empirical Methods in
  Natural Language Processing (EMNLP)}, pages 7222--7240, Online. Association
  for Computational Linguistics.

\bibitem[{Williams et~al.(2018)Williams, Nangia, and Bowman}]{mnli}
Adina Williams, Nikita Nangia, and Samuel Bowman. 2018.
\newblock \href {http://aclweb.org/anthology/N18-1101} {A broad-coverage
  challenge corpus for sentence understanding through inference}.
\newblock In \emph{Proceedings of the 2018 Conference of the North American
  Chapter of the Association for Computational Linguistics: Human Language
  Technologies, Volume 1 (Long Papers)}, pages 1112--1122. Association for
  Computational Linguistics.

\bibitem[{Wolf et~al.(2020)Wolf, Debut, Sanh, Chaumond, Delangue, Moi, Cistac,
  Rault, Louf, Funtowicz, Davison, Shleifer, von Platen, Ma, Jernite, Plu, Xu,
  Scao, Gugger, Drame, Lhoest, and Rush}]{transformers}
Thomas Wolf, Lysandre Debut, Victor Sanh, Julien Chaumond, Clement Delangue,
  Anthony Moi, Pierric Cistac, Tim Rault, Rémi Louf, Morgan Funtowicz, Joe
  Davison, Sam Shleifer, Patrick von Platen, Clara Ma, Yacine Jernite, Julien
  Plu, Canwen Xu, Teven~Le Scao, Sylvain Gugger, Mariama Drame, Quentin Lhoest,
  and Alexander~M. Rush. 2020.
\newblock \href {https://www.aclweb.org/anthology/2020.emnlp-demos.6}
  {Transformers: State-of-the-art natural language processing}.
\newblock In \emph{Proceedings of the 2020 Conference on Empirical Methods in
  Natural Language Processing: System Demonstrations}, pages 38--45, Online.
  Association for Computational Linguistics.

\bibitem[{Yanaka et~al.(2019{\natexlab{a}})Yanaka, Mineshima, Bekki, Inui,
  Sekine, Abzianidze, and Bos}]{yanakaMED}
Hitomi Yanaka, Koji Mineshima, Daisuke Bekki, Kentaro Inui, Satoshi Sekine,
  Lasha Abzianidze, and Johan Bos. 2019{\natexlab{a}}.
\newblock \href {https://doi.org/10.18653/v1/W19-4804} {Can neural networks
  understand monotonicity reasoning?}
\newblock In \emph{Proceedings of the 2019 ACL Workshop BlackboxNLP: Analyzing
  and Interpreting Neural Networks for NLP}, pages 31--40, Florence, Italy.
  Association for Computational Linguistics.

\bibitem[{Yanaka et~al.(2019{\natexlab{b}})Yanaka, Mineshima, Bekki, Inui,
  Sekine, Abzianidze, and Bos}]{yanakaHELP}
Hitomi Yanaka, Koji Mineshima, Daisuke Bekki, Kentaro Inui, Satoshi Sekine,
  Lasha Abzianidze, and Johan Bos. 2019{\natexlab{b}}.
\newblock \href {https://doi.org/10.18653/v1/S19-1027} {{HELP}: A dataset for
  identifying shortcomings of neural models in monotonicity reasoning}.
\newblock In \emph{Proceedings of the Eighth Joint Conference on Lexical and
  Computational Semantics (*{SEM} 2019)}, pages 250--255, Minneapolis,
  Minnesota. Association for Computational Linguistics.

\bibitem[{Zhang and Bowman(2018)}]{zhangbowman}
Kelly Zhang and Samuel Bowman. 2018.
\newblock \href {https://doi.org/10.18653/v1/W18-5448} {Language modeling
  teaches you more than translation does: Lessons learned through auxiliary
  syntactic task analysis}.
\newblock In \emph{Proceedings of the 2018 {EMNLP} Workshop {B}lackbox{NLP}:
  Analyzing and Interpreting Neural Networks for {NLP}}, pages 359--361,
  Brussels, Belgium. Association for Computational Linguistics.

\end{thebibliography}
\bibliographystyle{acl_natbib}
\end{document}